# Deep Learning for Large-Scale Real-World ACARS and ADS-B Radio Signal Classification

**Shichuan Chen, Shilian Zheng, (Member, IEEE), Lifeng Yang, and Xiaoniu Yang**
Science and Technology on Communication Information Security Control Laboratory, Jiaxing 314033 China

Corresponding author: Xiaoniu Yang (e-mail: yxn2117@126.com).

This work was supported in part by National Natural Science Foundation of China under Grant No. 61871398.

**ABSTRACT** Radio signal classification has a very wide range of applications in the field of wireless communications and electromagnetic spectrum management. In recent years, deep learning has been used to solve the problem of radio signal classification and has achieved good results. However, the radio signal data currently used is very limited in scale. In order to verify the performance of the deep learning-based radio signal classification on real-world radio signal data, in this paper we conduct experiments on large-scale real-world ACARS and ADS-B signal data with sample sizes of 900,000 and 13,000,000, respectively, and with categories of 3,143 and 5,157 respectively. We use the same Inception-Residual neural network model structure for ACARS signal classification and ADS-B signal classification to verify the ability of a single basic deep neural network model structure to process different types of radio signals, i.e., communication bursts in ACARS and pulse bursts in ADS-B. We build an experimental system for radio signal deep learning experiments. Experimental results show that the signal classification accuracy of ACARS and ADS-B is 98.1% and 96.3%, respectively. When the signal-to-noise ratio (with injected additive white Gaussian noise) is greater than 9 dB, the classification accuracy is greater than 92%. These experimental results validate the ability of deep learning to classify large-scale real-world radio signals. The results of the transfer learning experiment show that the model trained on large-scale ADS-B datasets is more conducive to the learning and training of new tasks than the model trained on small-scale datasets.

## I. INTRODUCTION

Radio signal classification has a very wide range of applications in the field of wireless communication and electromagnetic spectrum management [1]-[3]. In adaptive modulation and coding communication, the receiver can recognize the modulation and coding mode used by the transmitter, and then demodulate and decode the received signal by using corresponding demodulation and decoding algorithms, which helps to reduce the protocol overhead. In the field of spectrum management, cognitive radio [4] can detect the primary user signal by identifying the radio signal in the sensing band, thereby avoiding harmful interference to the primary user. Furthermore, by identifying various illegal users and interference signals, the security of the physical layer of the wireless communication can be improved, and the legal use of the spectrum can be guaranteed [5][6].

In recent years, with the development of deep learning [7][8] technology, it has been widely used in the fields of image recognition [9][10], speech recognition [11], natural language processing [12], and wireless communications [13]-[20]. In the past two years, deep learning has also been used to solve the problem of radio signal classification [21][22] and has obtained superior performance over traditional feature-based methods. However, the existing dataset used in the study of radio signal classification based on deep learning is relatively limited in scale, and many of the data are generated by the USRP. There is little research on the real-world signal data radiated by existing commercial radio transmitters. In order to verify the performance of the deep learning-based radio signal classification method on real radio signal data, we conduct experiments with large-scale real-world ACARS (Aircraft Communications Addressing and Reporting System) [23] and ADS-B (Automatic Dependent Surveillance - Broadcast) [24] signal data. Furthermore, we use the same convolutional neural network (CNN) model to realize ACARS signal classification and ADS-B signal classification, which verifies the ability of a single basic deep neural network model to deal with different types of radio signals (communication signal bursts and pulse bursts).

### A. RELATED WORK

The research on radio signal classification based on deep learning mainly focuses on two aspects: automatic modulation classification and radio frequency (RF) fingerprinting. In deep learning based automatic modulation classification, the authors used a simple CNN for modulation classification [25], and the experimental results show that the method obtained performance close to the feature-based expert system. They further exploited the deep residual (ResNet) network to improve the classification performance [21]. In addition to CNN and ResNet, the performance of six neural network models are compared in [26] and the training complexity is reduced from the perspective of reducing the input signal dimension. In most cases, the input size is fixed CNN-based modulation classification. The authors [27] proposed three fusion methods to improve the classification accuracy when the signal length is greater than the CNN input length. The above studies are all based on the classification of raw in-phase and quadrature (IQ) data. In addition to IQ, other forms of signal input are also considered. For example, the authors [28] used the instantaneous amplitude and instantaneous phase of the signal as the input of the long short-term memory (LSTM) network for modulation classification. Others convert IQ data into images by transformation, and then use the deep learning method of image classification to classify radio modulation [29][30]. In addition, there are many works employing generative adversarial networks (GANs) to analyze the security of the classification network [31] or for data augmentation [32]. In terms of modulation classification, the size of the signal dataset in [21] is relatively representative. The data set includes 24 modulation types, and the signals are transmitted and received through the USRP.

In addition to automatic modulation classification, RF fingerprinting has also begun to adopt deep learning methods. A CNN was used in [22] for fingerprinting identification of five ZigBee devices. Bispectrum of the received signal was calculated as a unique feature in [33] and a CCN was used to identify specific emitters. A CNN was also used in [34] to identify five USRP devices. The deep learning-based method obtained better performance than SVM and Logistic regression-based methods. 16 static USRP devices placed in a room were considered in [35] for CNN-based emitter classification. In order to consider a much more realistic where the topology evolving over time, a series of datasets gathered with 21 emitters were considered in [36] for emitter classification and state-of-the-art performance was obtained.

**TABLE 1. Radio signal data set used in some of the works.**

| Tasks | No. of categories | Data generation method |
|---|---|---|
| Modulation classification [21] | 24 | USRP |
| RF fingerprinting identification [22][36] | 7 (ZigBee), 21 (USRP) | USRP or commercial ZigBee |
| ACARS classification [2] | 2016 | Real-world ACARS transmitters |

In [2], we used CNN and the commercial real world ACARS signal data to identify the aircrafts. The number of aircraft classified is 2,016, and the total sample size is 60,480. Table 1 summarizes the data used in the current classification of radio signals based on deep learning. It can be seen that except the dataset used in our prior work [2], the radio signal dataset used in the study of radio signal classification based on deep learning is very limited in scale.

In order to further verify the performance of deep learning on the real-world large-scale radio signal classification, in this paper we expand the radio signal dataset scale, and carry out experimental verification from two datasets, ACARS and ADS-B, respectively. From the perspective of the sample size and the number of classification categories, the dataset used in this paper is the largest real-world dataset used for radio signal classification so far.

### B. CONTRIBUTIONS AND STRUCTURE OF THE PAPER

In summary, the contributions of the paper are as follows:
- We test the radio signal classification method based on deep learning on large-scale real-world datasets. The datasets used include ACARS and ADS-B. Large-scale is reflected in the sample size and the number of categories. Sample sizes of ACARS and ADS-B are as high as 900,000 and 13,000,000, respectively. The number of ACARS classification categories is 3,143, and the number of ADS-B classification categories is 5,157. To the best of our knowledge, it is the first time that a deep learning method for radio signal classification has been carried out on such large-scale real-world data.
- We use the same CNN model structure for both ACARS signal classification and ADS-B signal classification. The ACARS signal is in the form of a communication burst, while the ADS-B signal is in the form of a pulse burst. In this paper, the same CNN model is used for ACARS classification and ADS-B classification, which verifies the ability of a single basic deep neural network model to deal with different types of radio signals.
- We build an experimental system for radio signal deep learning experiments. The experimental system operates in the 30MHz-3GHz frequency band and supports large-scale radio signal acquisition, distribution, storage, training, and real-time reference. The classification experiments of ACARS and ADS-B signals in this paper are conducted on this experimental system.

The rest of this paper is organized as follows. In Section II, we introduce ACARS and ADS-B signals and the datasets used in the rest of the paper. In Section III, we present the CNN model. In Section IV we discuss the experimental results, and finally in Section V we summarize the paper.

## II. LARGE-SCALE REAL-WORLD RADIO SIGNAL DATASETS

### A. INTRODUCTION OF RADIO SIGNALS

There are many different types of radio signals. In order to facilitate data acquisition and analysis, the radio signals used in this paper are ACARS and ADA-B.

ACARS is a digital data link system that transmits short messages between aircrafts and ground stations via radio or satellite. The VHF ground-to-air data link with the ACARS system can realize real-time two-way communication of ground-air data with the transmission protocol ARINC618. This paper uses the downlink signal. Each ACARS message contains a maximum of 220 bytes. The data frame format is shown in Fig. 1.

ADS-B has been widely used as a next-generation air traffic management surveillance technology worldwide. An ADS-B message consists mainly of two parts, the preamble pulse part and the data pulse part, as shown in Fig. 2. The ADS-B considered in this paper is the 1090 MHz Extended Squitter (1090ES) mode. There is also an S-mode long response signal at 1090 MHz, which also conforms to the pulse specification shown in Fig. 2, but with specific data bit information. The ADS-B dataset in this paper contains both ADS-S signals and S-mode long response signals.

### B. THE LARGE-SCALE DATASETS

We use an acquisition system to receive and label the VHF band ACARS signals and the 1090 MHz ADS-B and S mode response signals. The system components are described in Section IV. After a long time of collection, some samples are selected, and the two datasets formed are as follows. Some of the samples are shown in Fig. 3.

- ACARS: The total sample size is 900,000. The number of categories (number of aircrafts) is 3,143, and the sample size of each category ranges from 60 to 1,000. The sample rate is 16 ksps and the length of each sample is 13,500. The labels are the aircraft IDs. 40 samples of each category are selected to form the test set, and the remaining samples are used as training samples. Three of these samples are shown in Fig. 3 (a).
- ADS-B: The total sample size is 13,000,000. The number of categories (number of aircrafts) is 5,157, and the sample size of each category varies from 150 to 9,400. The sample rate is 100 Msps and each sample is 13,500 in length. The labels are the aircraft IDs. For each category, 50 samples are selected to form the test set, and the remaining samples are used as training samples. Some of the samples are shown in Fig. 3 (b). It can be seen that some samples contain external interference pulses.

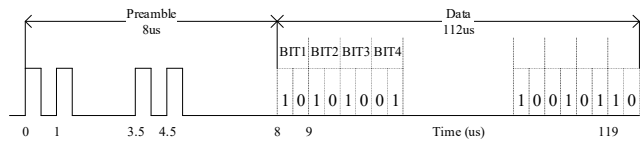

**FIGURE 1.** ACARS data frame format.

**FIGURE 2.** ADS-B pulse specification. The ADS-B considered in this paper is the 1090 MHz Extended Squitter (1090ES) mode.

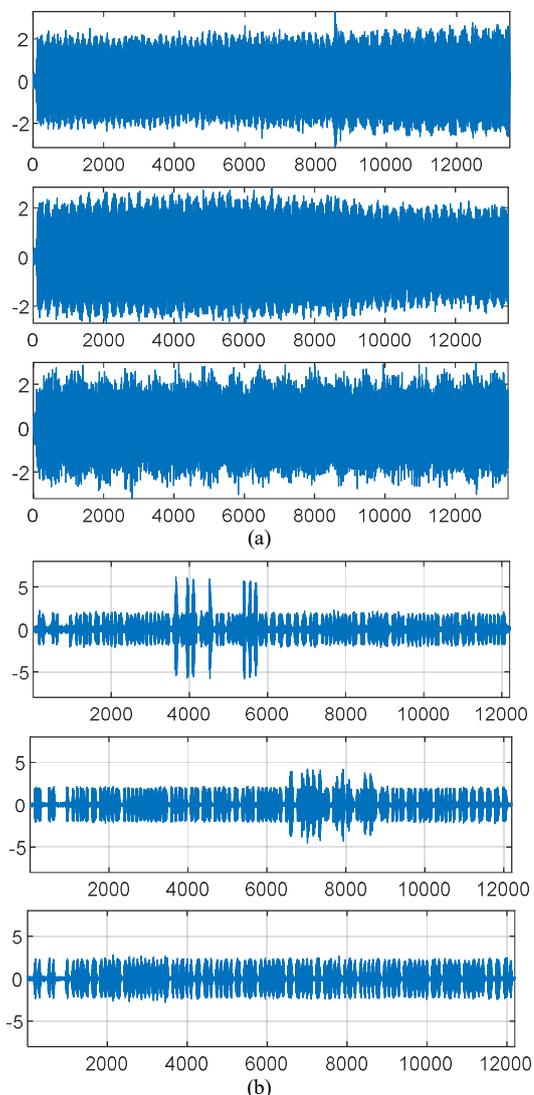

**FIGURE 3.** Examples of ACARS and ADS-B signals. (a) Three ACARS signal samples and (b) three ADS-B signal samples.

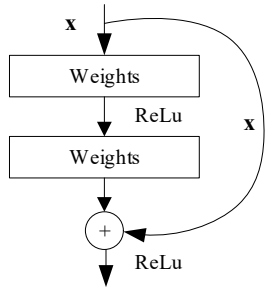

**FIGURE 4.** The residual module [37].

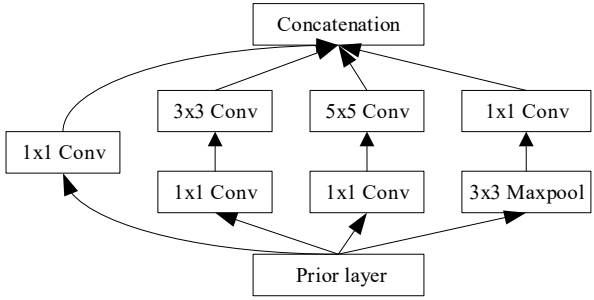

**FIGURE 5.** The Inception module [38].

## III. THE CNN MODEL

CNNs are a widely used deep neural network. Most CNN structures are inspired by LeNet. Classic CNNs often contain four basic layers: convolutional layer, normalized layer, nonlinear activation layer and pooling layer. In recent years, with the development of research, various special CNN models have emerged, including ResNet [37], Inception CNN [38], DenseNet [39], and so on.

The CNN structure considered in this paper is the Inception-residual network structure. As the depth of the traditional CNN increases, the training error will become larger, that is, the degradation problem occurs. In [37], the authors proposed a residual network to solve the training problem of deep networks. The basic component of the residual network is shown in Fig. 4.

The channels of each layer of the Inception network are provided with convolution kernels of different sizes, and then the respective channels are connected and combined to the next layer. Therefore, the Inception network is characterized by multi-resolution analysis. The basic module is shown in Fig. 5. Since there are many sizes of convolution kernels in each layer, the learning ability is improved, which is beneficial to improve the network performance.

Based on the Inception module and the residual module, we construct a deep Inception residual network structure for radio signal classification, as shown in Fig. 6. Details of the Inception-res blocks are given in Appendix. The network input is the original sample sequence of the received signal. In this paper we use this same model structure to achieve signal classification for both ACARS and ADS-B datasets.

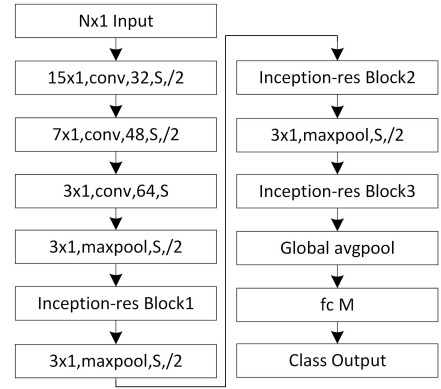

**FIGURE 6.** Deep Inception-residual network layout. In the figure, "conv" represents the convolutional layer; the number before "conv" represents the size of the convolution kernel, and the following number indicates the number of convolution kernels; "S" indicates that the convolution contains padding so that the input and output are of the same size, and "/2" indicates that the downsampling factor is 2, which means the output size is reduced to half of the input size; "maxpool" represents the maximum pooling; "Global avgpool" represents the global average pooling; "fc" represents the fully connected layer, and the number after that represents the number of neurons; "depth cat" indicates the concatenation layer; M is the number of categories; the final output is the category. All layers are activated by ReLU, and the batch normalization layer is also included between the convolutional layer and the nonlinear activation layer, which is not shown in the figure for the sake of simplicity. Details of Inception-res Block1, Inception-res Block2, and Inception-res Block3 are given in Appendix.

## IV. EXPERIMENTAL RESULTS

### A. EXPERIMENTAL SYSTEM

The experimental system is upgraded on our previous big data processing system [2] for radio signals. The physical composition is shown in Fig. 7. It is mainly composed of an antenna, an RF receiving and sampling module, a storage system, a computation system and a data exchange network. Details are as follows:

- Antenna. The antenna used is an omnidirectional antenna with vertical polarization, a typical gain of 0 dB, and an operating frequency range of 30 MHz to 3 GHz, which can completely cover the ACARS and ADS-B bands.

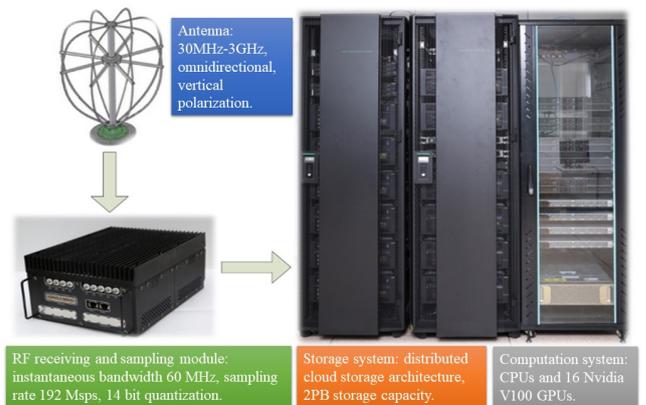

**FIGURE 7.** Experimental system.

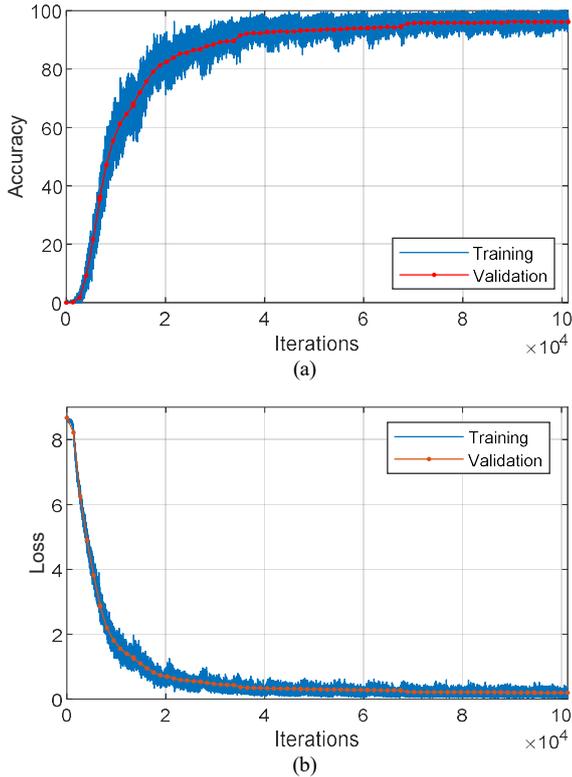

**FIGURE 8.** Training process of ADS-B classification. (a) accuracy and (b) loss.

- RF receiving and sampling module. The instantaneous bandwidth is 60 MHz, the ADC sampling rate is 192 Msps, and the quantization bit number is 14 bit.
- Storage system. The storage system stores the collected ACARS and ADS-B signals. With a distributed cloud storage architecture, the storage system has a capacity of 2 PB.
- Computation system. The computing system is mainly composed of CPUs and GPUs. The system consists of sixteen Nvidia V100 GPUs for deep learning training and reference tasks.
- Data exchange network. The data exchange network uses the InfiniBand (IB) network. It mainly distributes the signal sampling data obtained by the RF receiving and digitizing module to the storage system and the computing system, and realizes data exchange between the storage system and the computing system.

The reception, detection, labeling, training, and online classification of ACARS and ADS-B signals are implemented on the system. It should be noted that although we focus on ACARS and ADS-B radio signals in this paper, the system can also handle other radio signals in its operating frequency band.

### B. EXPERIMENTAL RESULTS

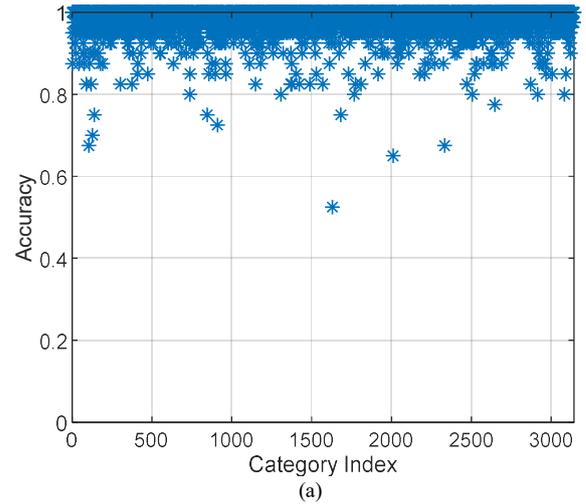
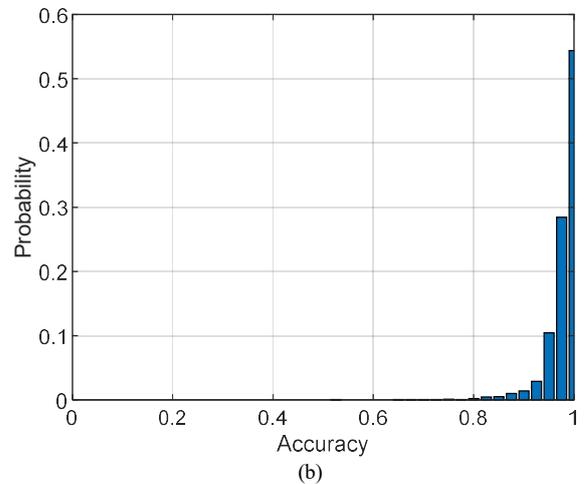

**FIGURE 9.** ACARS classification accuracy of the test set. (a) Classification accuracy of each category and (b) histogram of classification accuracy.

#### 1) TRAINING OF THE NETWORK
For the ACARS and ADS-B datasets, the Inception-Resnet model shown in Fig. 6 is used for classification experiments. Based on the training set and test set divided in Section II, we train the network model. We use the stochastic gradient descent (SGD) with momentum as the training method. The momentum factor is 0.9. The mini-batch size is set to 190. The maximum number of iterations for training is 101,250. Validation is performed every 1,350 iterations. Fig. 8 shows the training progress of ADS-B signal training. The training curves of the ACARS signals are similar and will not be given here for the sake of simplicity.

#### 2) BASIC CLASSIFICATION RESULTS
The test accuracy of ACARS data and ADS-B data is 98.1% and 96.3%, respectively. For ACARS, the classification accuracy of 4,863 in the 5,157 categories is greater than 90%. For ADS-B, the classification accuracy of 3,022 in the 3,143 categories is greater than 90%. Fig. 9 and Fig. 10 show the classification accuracy of each category of ACARS and ADS-B, respectively. It can be seen that the classification accuracy of a few categories is much lower than that of others.

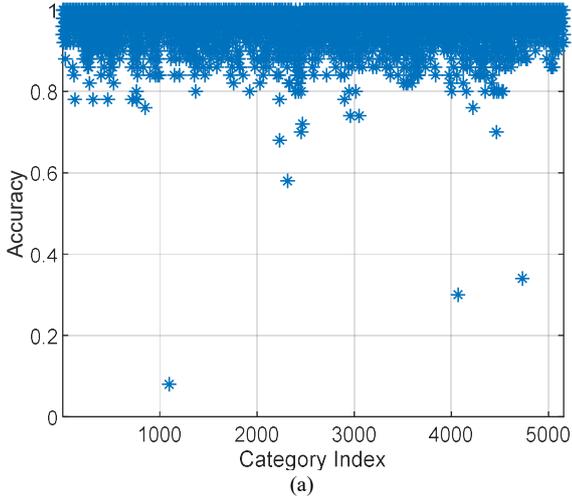

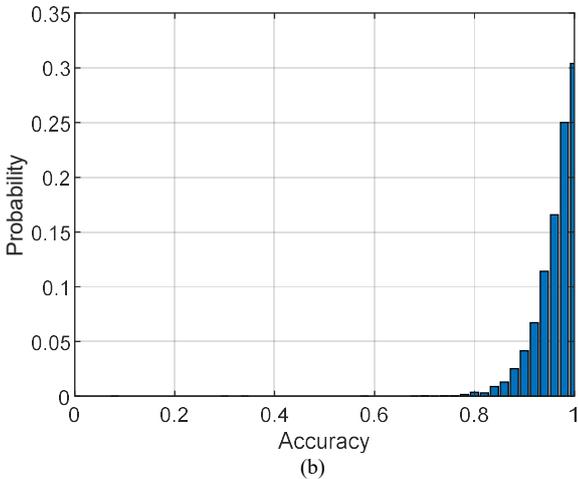

**FIGURE 10.** ADS-B classification accuracy of the test set. (a) Classification accuracy of each category and (b) histogram of classification accuracy.

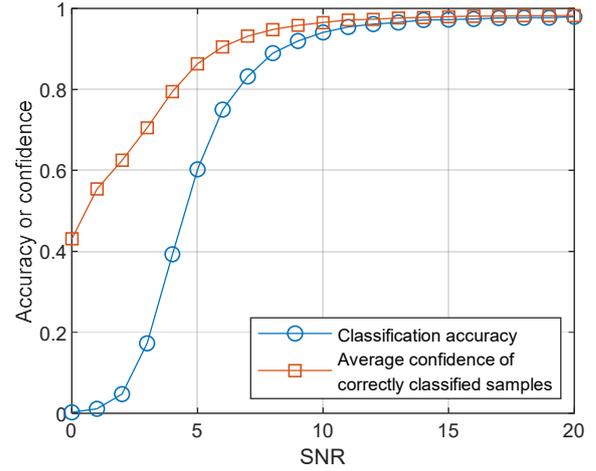

**FIGURE 11.** ACARS classification accuracy of the test set under different noise levels.

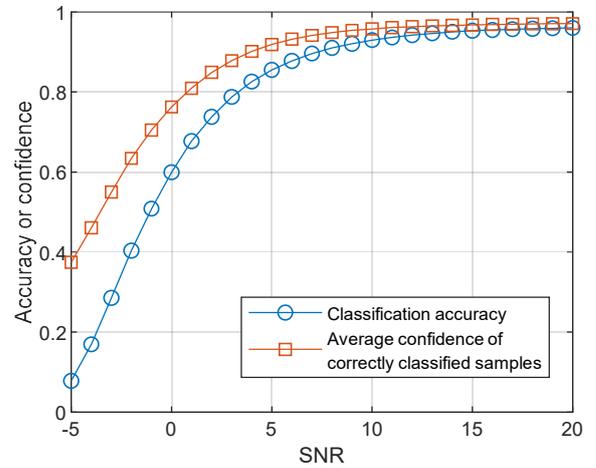

**FIGURE 12.** ADS-B classification accuracy of the test set under different noise levels.

### 3) PERFORMANCE WITH INJECTED WHITE GAUSSIAN NOISE

To measure the classification performance at different noise levels, we added additive white Gaussian noise to the test data for experiments. Fig. 11 and Fig. 12 show the classification performance of ACARS and ADS-B at different noise levels. It should be noted that the signal-to-noise ratio (SNR) here refers to the SNR after noise is added when we regard the original signal as a pure signal without noise (in fact, the original signal itself also contains noise). Therefore, the actual SNRs are smaller than those shown in the figure. It can be seen that although these noisy samples have not been trained, the model has good performance to classify these noisy samples. When the SNR is greater than 9 dB, the classification accuracy under both datasets is greater than 92%. The figures also illustrate the average confidence of the correctly classified samples under different SNRs. It can be seen that the lower the SNR, the lower the confidence. In addition, comparing the classification accuracy of ACARS and ADS-B at different SNRs, the ADS-B classification is more robust to noise, which may be caused by the ADS-B signal being a burst of pulses.

### 4) PERFORMANCE OF TRANSFER LEARNING

To further illustrate the importance of large-scale data deep learning, we also conducted a transfer learning (TL) experiment using the ADS-B classification as an example. Consider a new learning task that categorizes 190 categories of ADS-B signals. The signal data of these 190 categories are not in the dataset introduced in Section II. The number of samples per category is between 50 and 100. The total number of samples is 14,000. For performance comparison, we selected data of 50 and 500 categories from the dataset discussed in Section II, respectively. The total number of samples was 360,000 and 2,850,000, respectively. The trained networks with these two datasets are indicated as Net50 and Net500. The network trained with the entire dataset is indicated as NetAll. We conduct the training in two ways: without TL and with TL. The non-TL method directly trains with the data of 190 categories of ADS-B signals, while the TL method performs fine tuning on the three network models that have been trained, i.e., Net50, Net500, and NetAll. Fig. 13 shows the experimental results. It can be seen that TL method converges faster than non-TL training and obtains higher final classification accuracy. Specifically, the final classification accuracy of the non-TL method is 81.24%, and the classification accuracy rates of the three TL methods are 88.56%, 90.42%, and 92.55%, respectively. Table 2 further shows the number of training iterations required to achieve the target classification accuracy. It can be seen that as the size of data used to train the TL network increases, the time required for TL training is decreases. Especially with NetAll TL, in order to achieve 80% classification accuracy, the number of training iterations is reduced by more than 80 times. These indicate that the larger the dataset size, the more the network obtained by training with these data can support the training of new tasks, thus indicating the importance of large-scale datasets.

**TABLE 2. Number of iterations in training.**

| Target accuracy | | 60% | 70% | 80% | 90% |
|---|---|---|---|---|---|
| No. of iterations | Without TL | 520 | 650 | 1150 | - |
| | Net50 TL | 72 | 107 | 169 | - |
| | Net500 TL | 22 | 32 | 60 | 600 |
| | NetAll TL | 9 | 11 | 13 | 46 |

## V. CONCLUSION

We have used large-scale real-world ACARS and ADS-B radio signal data to verify the performance of deep learning-based radio signal classification. We have built an experimental system to conduct the experiments. The experimental results show that the signal classification accuracy of ACARS and ADS-B is 98.1% and 96.3%, respectively. When the signal-to-noise ratio (with injected additive white Gaussian noise) is greater than 9 dB, the classification accuracy is greater than 92%. The results of the transfer learning experiment show that the model trained on large-scale ADS-B datasets is more conducive to the learning and training of new tasks than the model trained on small-scale datasets. We have used the same Inception-Residual neural network model structure for ACARS signal classification and ADS-B signal classification to verify the ability of a single basic deep neural network model structure to process different types of radio signals. These results have verified the ability of deep learning for large-scale real-world radio signal classification. However, it can be seen from the current experimental results that the classification accuracy of some categories is significantly lower than other categories. Future research may consider resolving the problem through techniques such as ensemble learning.

## APPENDIX

Fig. 6 shows the basic CNN framework used in this paper. The structures of the three Inception-res blocks are shown in Fig. 14, Fig. 15 and Fig. 16.

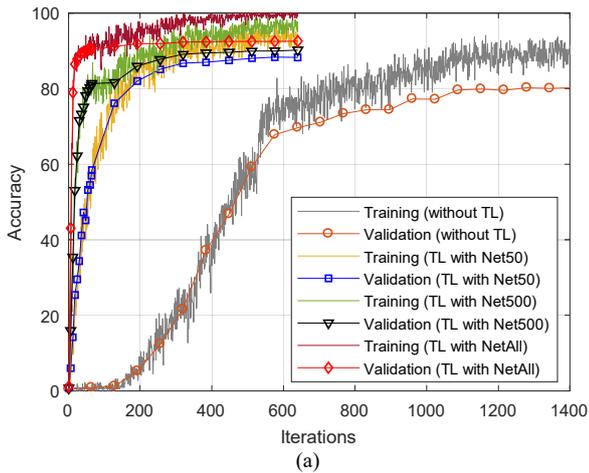
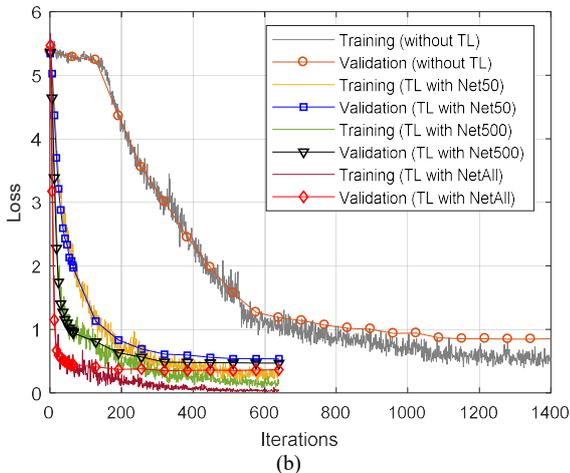

**FIGURE 13. Transfer learning of ADS-B classification.**

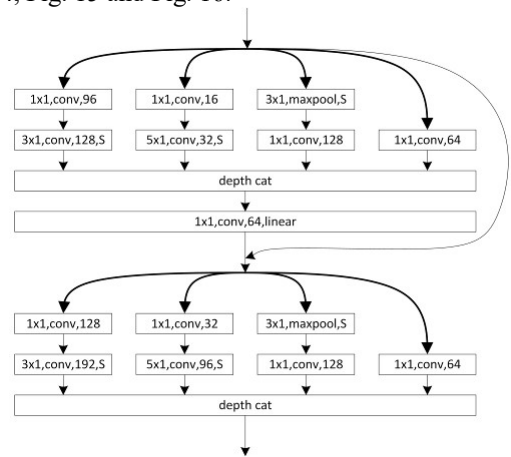

**FIGURE 14. Structure of Inception-res Block1 in Fig. 6.**

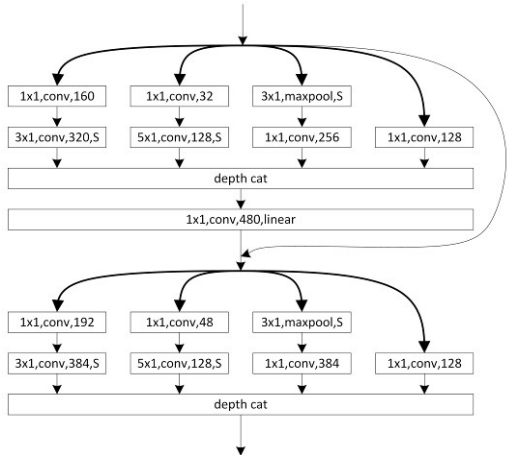

**FIGURE 15.** Structure of Inception-res Block2 in Fig. 6.

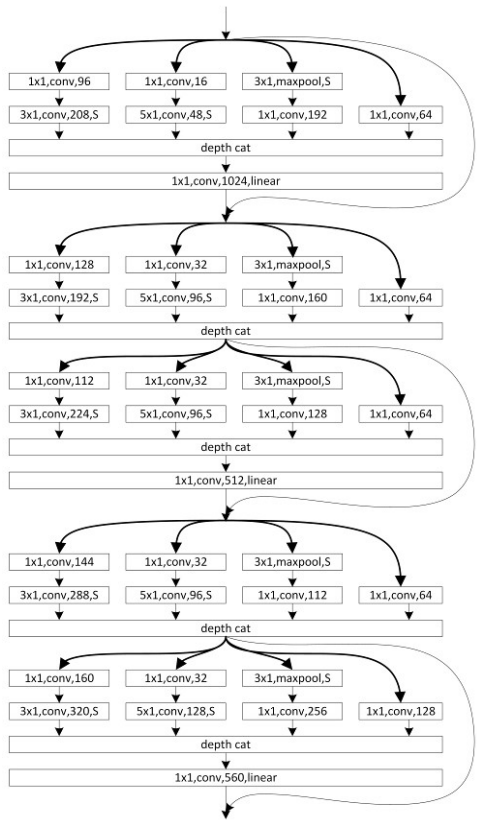

**FIGURE 16.** Structure of Inception-res Block3 in Fig. 6.